\documentclass[11pt]{article}
\usepackage[T1]{fontenc}
\usepackage{graphicx,verbatim}
\usepackage{booktabs}
\usepackage{amsmath}
\usepackage{subcaption} 
\usepackage{color}
\usepackage{multirow}
\usepackage{cleveref}
\usepackage{url}
\usepackage{amssymb}
\definecolor{rebuttal}{RGB}{255, 0, 0}
\begin{document}

\title{SAMSA: Segment Anything Model Enhanced with Spectral Angles for Hyperspectral Interactive Medical Image Segmentation}

\author{
  Alfie Roddan\textsuperscript{1} \\
  \and Tobias Czempiel\textsuperscript{1} \\
  \and Chi Xu\textsuperscript{1} \\
  \and Daniel S. Elson\textsuperscript{1} \\
  \and Stamatia Giannarou\textsuperscript{1}
}

\date{
  \textsuperscript{1}The Hamlyn Centre for Robotic Surgery, Imperial College London, UK\\[1ex]
  \today
}

\maketitle

\begin{abstract}
Hyperspectral imaging (HSI) provides rich spectral information for medical imaging, yet encounters significant challenges due to data limitations and hardware variations. We introduce SAMSA, a novel interactive segmentation framework that combines an RGB foundation model with spectral analysis. SAMSA efficiently utilizes user clicks to guide both RGB segmentation and spectral similarity computations. The method addresses key limitations in HSI segmentation through a unique spectral feature fusion strategy that operates independently of spectral band count and resolution. Performance evaluation on publicly available datasets has shown 81.0\% 1-click and 93.4\% 5-click DICE on a neurosurgical and 81.1\% 1-click and 89.2\% 5-click DICE on an intraoperative porcine hyperspectral dataset. Experimental results demonstrate SAMSA's effectiveness in few-shot and zero-shot learning scenarios and using minimal training examples. Our approach enables seamless integration of datasets with different spectral characteristics, providing a flexible framework for hyperspectral medical image analysis.


\end{abstract}

\section{Introduction}

Hyperspectral imaging (HSI) offers superior intraoperative guidance through its rich spectral information, allowing precise differentiation between visually similar tissues \cite{Clancy2020,TOM201800455}. 
The diverse range of HSI hardware, with varying spectral ranges and resolutions, creates significant interoperability challenges that impede data standardization \cite{ANICHINI2024108293}. This technical fragmentation, coupled with HSI's limited clinical adoption, has resulted in a shortage of comprehensive datasets, presenting a substantial obstacle for machine learning applications \cite{Goodfellow-et-al-2016}. Despite these challenges, recent advances have demonstrated HSI's potential for intraoperative segmentation in neurosurgery \cite{Shapey2019,roddan2024,Leon2023} and on porcine organs \cite{SEIDLITZ2022102488,Studier-Fischer2023}. However, developing generalized models that account for hardware variations remains unsolved.
Classical Spectral Comparison Functions (SCF) such as Spectral Angle (SA) \cite{boardman1993spectral} and Pearson's Correlation Coefficient (PCC) \cite{Meneses2000SpectralCM} offer highly adaptable approaches for comparing spectra for manual image segmentation. Their untrained nature allows them to generalize to new scenarios without requiring additional data, functioning with any spectral range or number of bands. These methods typically operate by using a reference point (user click) to compare against the rest of the image, making them inherently interactive. However, they face limitations due to the "shading problem" where semantic objects exhibit different spectral signatures, and the challenge of establishing consistent segmentation thresholds within and across images.

Interactive segmentation is particularly valuable in medical imaging, as it leverages expert input to improve performance compared to fully automated methods \cite{zhao2013overview,Wang2018,Wang_2019} and enables segmentation of previously unseen tissue classes — a vital capability during surgical procedures where unexpected pathological findings may occur. The shared interactive nature of both classical spectral methods and modern RGB interactive segmentation presents a natural opportunity to combine these approaches, allowing a single user click to serve dual purposes namely, guiding the RGB-based model while simultaneously providing a reference point for spectral comparison. While powerful interactive models like Segment Anything and its successor SAM2 \cite{kirillov2023segment,ravi2024sam2segmentimages} have revolutionized RGB segmentation, these advances cannot be directly applied to HSI due to fundamental differences in data characteristics. 

In this work, we propose an interactive image segmentation approach by combining SAM2 with spectral analysis techniques to overcome HSI's data limitations. Our approach leverages the advantages of large-scale RGB foundation models and integrating HSI's rich spectral information. 
Specifically, we contribute: (1) An interactive segmentation framework for HSI, utilizing a dual-input approach that efficiently leverages the same user input (clicks) in two complementary ways: to guide an RGB foundation model and to compute SCF measurements in HSI data, enhancing segmentation performance. (2) We demonstrate effectiveness in both few-shot and zero-shot learning scenarios for tumor classification, showing robust performance even with extremely limited training examples and on unseen test cases. (3) The first HSI machine learning framework that functions independently of HSI band count and wavelength variations, enabling the combination of datasets with different spectral characteristics and semantic classes into a unified training approach.

\section{Methodology}
\begin{figure}
\includegraphics[width=\textwidth]{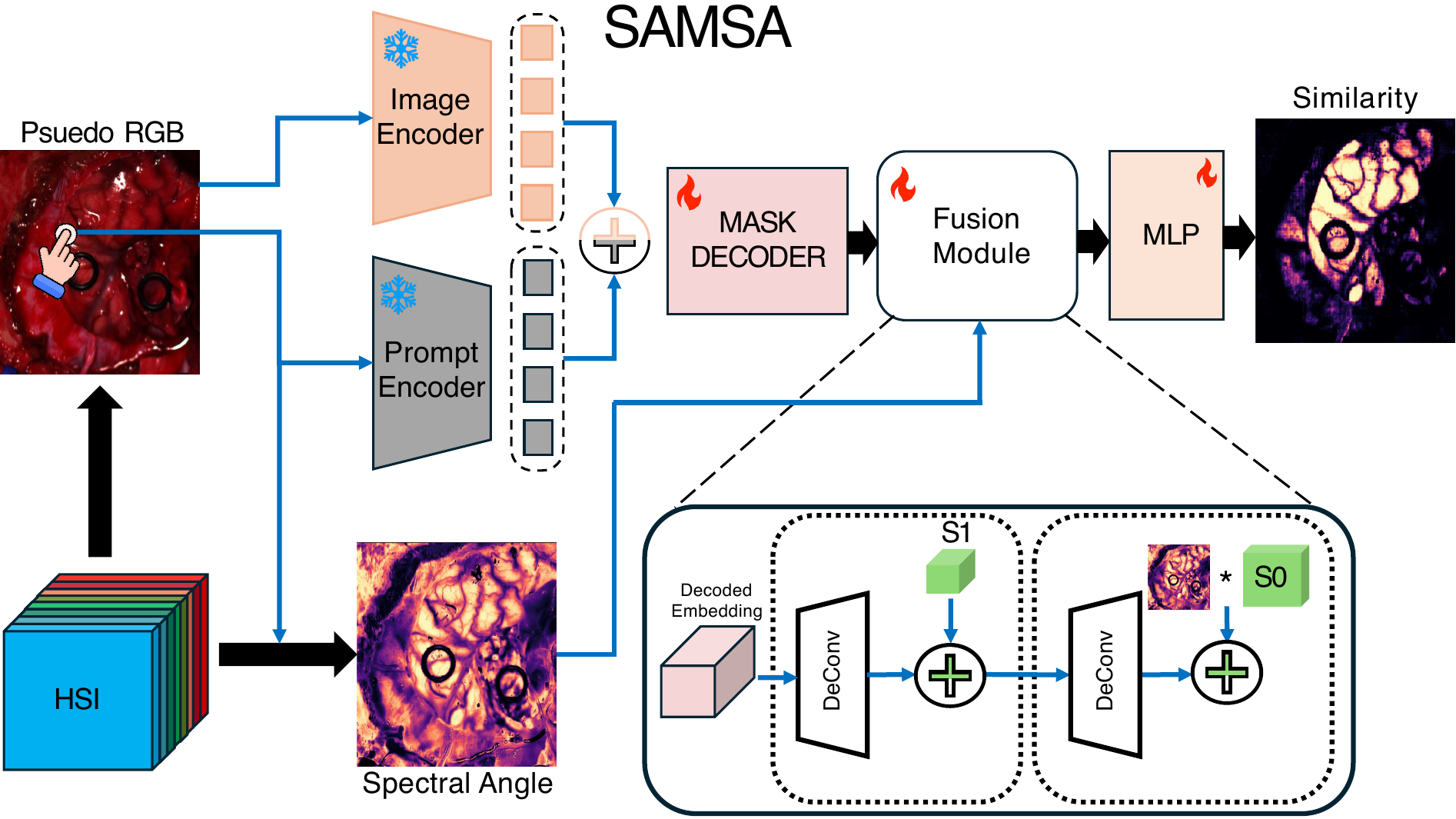}
\caption{SAMSA outline - a single click in the pseudo RGB is used to guide both the RGB and spectral branch.} \label{model_outline}
\end{figure}

Given a hyperspectral image $X \in \mathbb{R}^{H \times W \times C}$, where $H$ and $W$ denote the spatial dimensions and $C$ represents the number of spectral channels, our goal is to perform interactive foreground/background segmentation based on user-provided click positions. Additionally, we have available a corresponding pseudo RGB image $X_{rgb} \in \mathbb{R}^{H \times W \times 3}$ and a ground truth label map $Y \in {[0,...,N]}^{H \times W}$ with $N$ the number of classes. Let $\mathcal{I} = \{I_{i,j}\}$ be a set of user-provided click positions, where each $I_{i,j} = (i, j)$ corresponds to a pixel location in the image. Each model outputs a foreground/background similarity map $\hat{Y} \in [0,1]^{H \times W}$ with $0$ representing no similarity and  $1$ representing strong similarity to the clicked pixel(s).

\textbf{Spectral Comparison Function (SCF)}.
To model the spectral characteristics, we employ a spectral similarity approach based on click positions. 
Given multiple clicks $\mathcal{I}$ in the region of interest, we compute the similarity for each pixel in the image with respect to all selected spectra $S_{i,j} = X(I_{i,j})$ and assign the highest similarity score. Finally, the SCF outputs a similarity map $\hat{Y}_{SCF}=SCF(X, \mathcal{I})$. SA \cite{boardman1993spectral} measures the similarity between two spectra using the angle between them in the spectral space. The spectra can also be compared using PCC \cite{Meneses2000SpectralCM} which quantifies the linear relationship between the reference spectra and candidate spectra. PCC specifically focuses on modeling negative correlations between spectra distinguishing between positive and negative relationships. While SA and PCC effectively measure similarity between spectral samples, such as those derived from click data, they do not establish decision boundaries since they are not based on learned models. To address this, we employ histogram equalization to maximize the information content of the similarity maps by increasing contrast and improving regional separability \cite{gonzalez2018digital}. We denote this method as $SCF_{Equalized}$.










\textbf{RGB Interactive Segmentation}.
To leverage powerful RGB foundational models for HSI, we first generate pseudo RGB images from the HSI data through spectral band selection and combination~\cite{czempiel2024}. SAM2 \cite{ravi2024sam2segmentimages} is utilized as our RGB segmentation backbone due to its state-of-the-art performance in interactive segmentation tasks and its robust generalization capabilities across diverse imaging domains, including medical \cite{murali2024cyclesamoneshotsurgicalscene}. SAM2 generates confidence maps $\hat{Y}_{SAM2} = \text{SAM2}(X_{rgb}, \mathcal{I})$ indicating the likelihood of each pixel belonging to the foreground from the pseudo RGB image. $\text{SAM2}_{Base}(\cdot)$ denotes the SAM2 model with frozen weights, and $\text{SAM2}_{Tuned}(\cdot)$ denotes the fine-tuned version.


\textbf{RGB and Spectral Similarity}.
Both spectral and RGB-based models offer complementary information for image segmentation. While SAM2 processes only RGB information, the combination complements our spectral similarity-based segmentation by capturing spatial and contextual features that may not be evident in pure spectral analysis.
To enhance segmentation quality, we explored two initial approaches for fusing the spectral and spatial similarity maps. The first is a simple intersection method where the similarity maps are directly multiplied: $\hat{Y}_{\text{SAM2}_{Intersec.}} = \hat{Y}_{SAM2} \cdot \hat{Y}_{SCF}$. This multiplication produces high values only in regions where both modalities agree, effectively creating a logical AND operation that requires consensus between spectral and spatial information for pixel classification. For a more sophisticated integration, we implement a UNet architecture \cite{ronneberger2015unetconvolutionalnetworksbiomedical} that takes the similarity maps $\hat{Y}_{SAM2}$ and $\hat{Y}_{SCF}$ as direct inputs to learn optimal fusion strategies: $\hat{Y}_{\text{SAM2}_{UNet}} = \text{SAM2}_{UNet}(\hat{Y}_{SAM2}, \hat{Y}_{SCF})$, where $\text{SAM2}_{UNet}(\cdot)$ represents the trained fusion UNet model. Unlike the deterministic intersection approach, this learnable fusion should uncover complementary spatial relationships between modalities.



\textbf{SAMSA}.
To further improve segmentation, we introduce SAMSA, a novel model that fuses spectral similarity with high-resolution spatial features from SAM2. Unlike the aforementioned fusion approaches that combine outputs after segmentation, SAMSA integrates spectral information directly into the upscaling process of the SAM2 mask decoder. A high-level overview of this process is shown in Fig. \ref{model_outline}. Given $X_{RGB}$ and $\mathcal{I}$, SAMSA follows the standard SAM2 processing pipeline and additionally integrates the spectral information. The spectral similarity map $\hat{Y}_{SCF}$ is fused with the high-resolution feature maps $S_0$ extracted from SAM2's encoder, enhancing segmentation decisions based on spectral properties. This allows the model to leverage spectral characteristics that are not visible in pseudo RGB while maintaining SAM2's spatial precision. We freeze the prompt and image encoders from SAM2, fine-tuning only the lightweight mask decoder. This enables SAMSA to generalize to medical datasets with minimal training data while learning how to effectively combine spatial and spectral information.

\section{Experimental Results}
For training of the fusion models we mainly follow SAM's optimization procedure \cite{kirillov2023segment}. All models are trained with a combined loss function using DICE and cross-entropy loss with equal weighting, excluding any unlabeled regions. Complete implementation details are provided in the accompanying source code, accessible upon acceptance of this manuscript\footnote{\url{REDACTED_CODE_REPOSITORY_LINK}}.


\textbf{Datasets}. The \textbf{HiB dataset} includes hyperspectral and pseudo RGB images from 34 patients, with patient-wise fold splits~\cite{Leon2023}. It features four labeled classes: Background, Tumor, Healthy, and Vasculature, plus an Unlabeled category. Following preprocessing as in \cite{martinez2019most}, the dataset consists of 128 spectral bands. The \textbf{HeiPorSPECTRAL (Heipor)} dataset, collected from 20 porcine subjects at Heidelberg University Hospital, provides HSI data with annotations for 20 distinct organs. Spectral information ranges from 500 nm to 1000 nm, and corresponding RGB images are derived from the HSI data \cite{Studier-Fischer2023}.

\textbf{Evaluation Protocol} We evaluate each model on foreground/background segmentation using a single user click, following SAM2's evaluation procedure \cite{kirillov2023segment}. For each class, we select a click position at the center of the largest connected component in the foreground region to avoid boundary ambiguity. We report two key metrics: \textbf{D@0.5} - The DICE score \cite{Srensen1948AMO} using the standard decision boundary of $0.5$. \textbf{D@Max} - The max DICE across all thresholds, representing optimal performance without predefined decision boundaries.

We report macro-averaged (Macro) and per-class results. We also evaluate multi-click performance by placing subsequent clicks on the target foreground class. Finally, for trainable models, we conduct N-shot evaluations (1, 3, 5, 10, and 20 examples) to analyze the relationship between training data availability and segmentation quality.

\begin{table} [t]
\centering
\caption{Macro Results: Performance of models with varying input modalities (Mod.) and user clicks. Bold indicates peak performance per metric and dataset. }\label{macro_table}
\begin{tabular}{l|l|cc|c|cc|c}
\multirow{3}{*}{Mod.} & \multirow{3}{*}{Model} & \multicolumn{3}{c|}{Heipor} & \multicolumn{3}{c}{Hib Dataset} \\
\cline{3-8}
 &  & \multicolumn{2}{c|}{1 click} & 5 clicks & \multicolumn{2}{c|}{1 click} & 5 clicks \\
\cline{3-8}
 &  & D@0.5 & D@Max & D@0.5 & D@0.5 & D@Max & D@0.5 \\
\hline
\multirow{3}{*}{HSI} 
 & PCC & 0.122 & 0.472 & 0.117 & $0.373^{\pm 0.019}$ & $0.885^{\pm 0.034}$ & $0.375^{\pm 0.018}$ \\
 & SA & 0.117 & 0.489 & 0.117 & $0.374^{\pm 0.019}$ & $0.889^{\pm 0.035}$ & $0.374^{\pm 0.019}$ \\
 & $\text{SA}_{Equalized}$ & 0.205 & 0.487 & 0.137 & $0.568^{\pm 0.038}$ & $0.885^{\pm 0.034}$ & $0.482^{\pm 0.033}$ \\
 \hline
\multirow{2}{*}{RGB} 
 & $\text{SAM2}_{Base}$ & 0.600 & 0.773 & 0.643 & $0.523^{\pm 0.056}$ & $0.727^{\pm 0.043}$ & $0.591^{\pm 0.069}$ \\
 & $\text{SAM2}_{Tuned}$ & 0.806 & \textbf{0.864} & 0.886 & $0.771^{\pm 0.059}$ & $0.905^{\pm 0.036}$ & $0.912^{\pm 0.025}$ \\
\hline
\multirow{3}{*}{Fusion} 
 & $\text{SAM2SA}_{Intersec.}$ & 0.634 & 0.755 & 0.647 & $0.605^{\pm 0.048}$ & $0.832^{\pm 0.033}$ & $0.674^{\pm 0.083}$ \\
 & $\text{SAM2SA}_{UNet}$ & 0.692 & 0.798 & 0.771 & $0.650^{\pm 0.115}$ & $0.778^{\pm 0.123}$ & $0.673^{\pm 0.096}$ \\
 & SAMSA (ours) & \textbf{0.811} & 0.863 & \textbf{0.892} & $\textbf{0.810}^{\pm 0.050}$ & $\textbf{0.929}^{\pm 0.028}$ & $\textbf{0.934}^{\pm 0.031}$ \\

\end{tabular}
\end{table}

In our evaluation of spectral similarity functions, SA outperformed PCC with improvements of $+0.017$ on Heipor and $+0.004$ on Hib datasets when measured by D@Max (\cref{macro_table}). We further enhanced SA with equalization ($SA_{Equalized}$), improving contrast around the 0.5 threshold to better align with RGB models, and adopted this as our spectral analysis method for subsequent experiments.

For RGB-only performance (\cref{macro_table}), $\text{SAM2}_{Base}$ demonstrated reasonable generalization to medical domains, achieving $0.600$ Macro D@0.5 on Heipor. However, \cref{class_table} reveals significant weaknesses on the Hib dataset's Vascular class ($0.335$), indicating limited generalization to domain-specific medical structures. Fine-tuning substantially improved performance, with $\text{SAM2}_{Tuned}$ achieving $0.757$ on Vascular and $0.869$ on Background classes.

Our analysis of fusion strategies revealed that late fusion approaches namely, $\text{SAM2SA}_{Intersec.}$ and $\text{SAM2SA}_{UNet}$, underperformed compared to $\text{SAM2}_{Tuned}$, though they improved upon $\text{SAM2}_{Base}$. This suggests spectral information requires earlier integration to enhance segmentation performance, which we implemented in SAMSA.

\begin{figure}[t]
\centering
\includegraphics[width=1.0\textwidth]{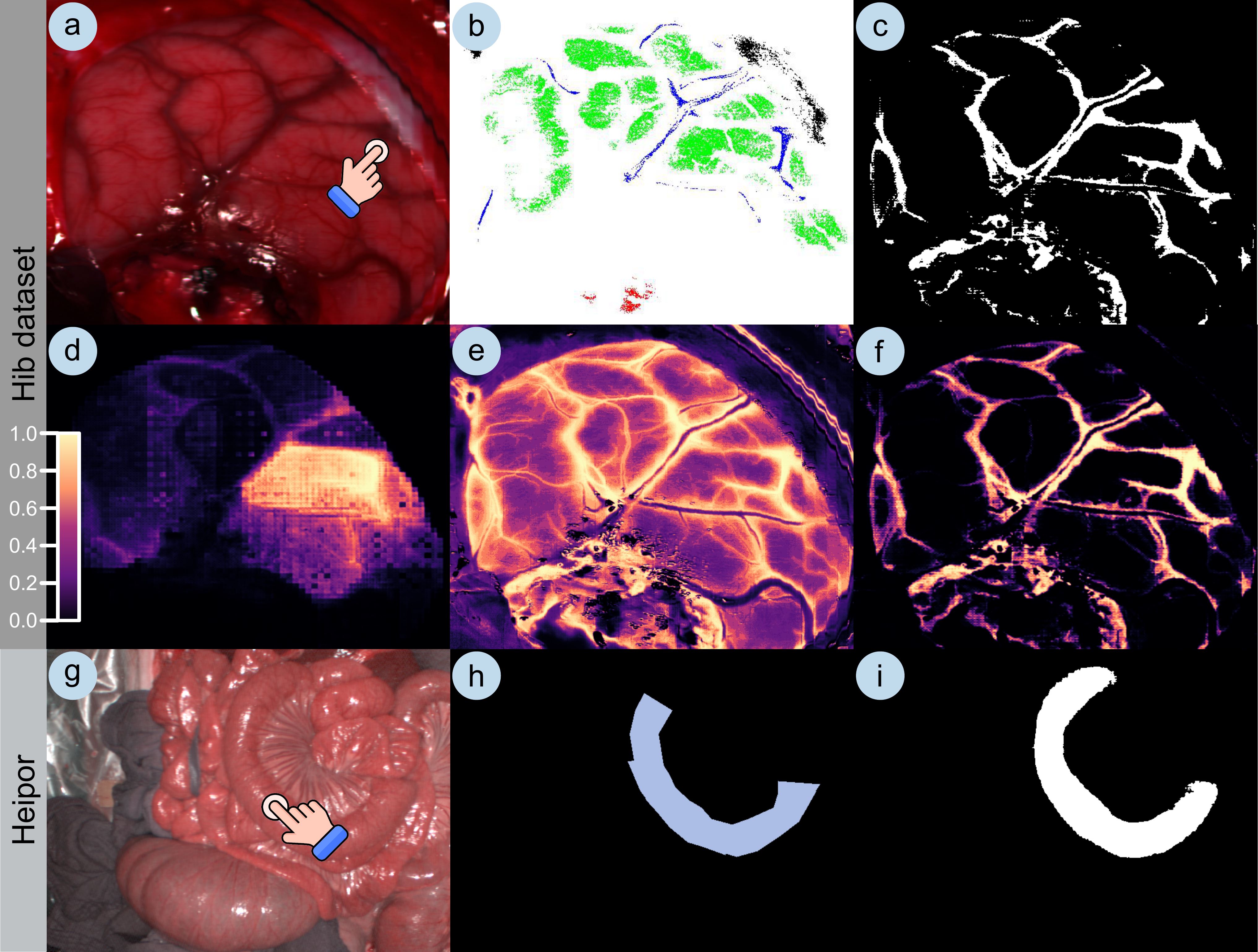}
\caption{Comparison of results. (a) RGB image with vascular click. (b) Corresponding label image, where Tumour is red, Vascular structures are blue, Healthy tissue is green, Background non-tissue structures are black, and Unlabeled regions are white. (c) SAMSA prediction. (d-f) Probability maps from SAM2, SA, and SAMSA. (g) RGB image with a small bowel click. (h) Corresponding label image, where Small Bowel is gray and Background is black. (i) SAMSA prediction.} \label{example_outputs}
\end{figure}

SAMSA consistently outperformed $\text{SAM2}_{Tuned}$ across all classes on D@0.5, with notable improvements of $+0.056$ for Healthy and $+0.06$ for Tumor classes. Macro D@0.5 scores increased by $+0.039$ for Hib and $+0.005$ for Heipor. The modest gains on Heipor can be attributed to its RGB-oriented annotations and predominance of large, centered objects (\cref{example_outputs}). These characteristics are particularly favorable for RGB-only models that detect visual boundaries, as evidenced by the strong zero-shot performance of $\text{SAM2}_{Base}$ ($0.773$), which trails the fine-tuned version by only $-0.091$ D@0.5. For this reason, we focused our per-class metric analysis on the Hib dataset, where spectral information provides more substantial benefits for segmentation.

\begin{figure}[t]
\centering
\includegraphics[width=0.95\textwidth]{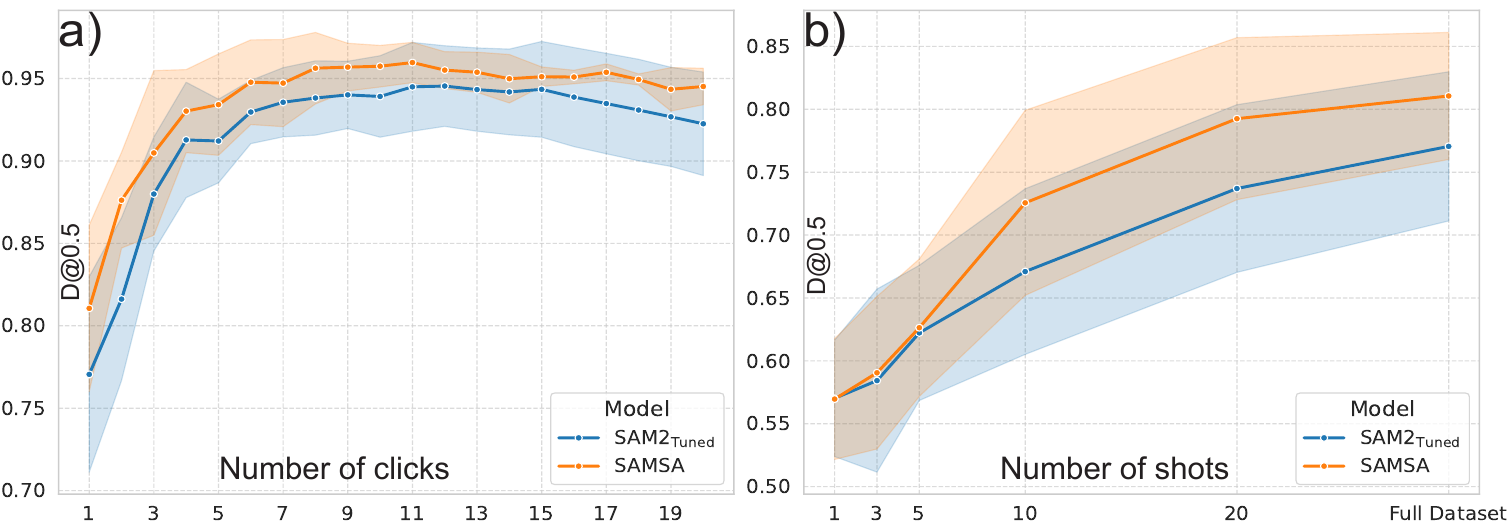}
\caption{Performance analysis on Hib dataset: a) Number of clicks and b) Number of shots in training and correlation to model performance.} \label{fig:performance_analysis}
\end{figure}

As expected, additional clicks improved segmentation performance for all fine-tuned models. SAMSA showed significant improvements with 5-click inputs, increasing performance of D@0.5 by $+0.081$ on Heipor and $+0.124$ on Hib. In \cref{fig:performance_analysis} we demonstrate SAMSA's superiority over $\text{SAM2}_{Tuned}$ across different click counts on Hib, achieving $0.95$ Macro D@0.5 with 5 clicks. Furthermore, with only 20 training examples, SAMSA achieves $0.79$ Macro D@0.5 for single-click segmentation. Leveraging foundation models, both SAMSA and SAM2 perform well in limited-data scenarios. Notably, the integration of spectral information consistently enhances the training process, with a clear performance gap between SAMSA and $\text{SAM2}_{Tuned}$ emerging at just 5 training examples, highlighting the advantage of spectral information in low-data regimes.

\textbf{Generalization Results}.
We conduct a leave-one-class-out experiment on both fine-tuned SAM2 and SAMSA by removing the Tumor class from training while testing across all classes on Hib, simulating real-world scenarios requiring identification of novel structures without prior supervision. 

\begin{table}[t]
\centering
\caption{Class results $D@0.5$ for Hib dataset using 1 click.}\label{class_table}
\begin{tabular}{l|ccccc}
 Model            & Macro   & Background    & Healthy       & Vascular      & Tumor        \\
\hline
 $SA_{Equalized}$  &  $0.568^{\pm 0.038}$  & $0.613^{\pm 0.109}$ & $0.815^{\pm 0.093}$ & $0.506^{\pm 0.130}$ & $0.339^{\pm 0.100}$ \\
 $\text{SAM2}_{Base} $     &     $0.523^{\pm 0.056}$     & $0.552^{\pm 0.079}$ & $0.586^{\pm 0.083}$ & $0.335^{\pm 0.095}$ & $0.619^{\pm 0.188}$ \\
$\text{SAM2}_{Tuned}$     &  $0.771^{\pm 0.059}$ & $0.869^{\pm 0.045}$ & $0.778^{\pm 0.081}$ & $0.757^{\pm 0.106}$ & $0.678^{\pm 0.098}$ \\
 SAMSA(ours) & $\textbf{0.810}^{\pm 0.050}$ & $\textbf{0.881}^{\pm 0.039}$ & $\textbf{0.834}^{\pm 0.066}$  & $\textbf{0.790}^{\pm 0.117}$ & $\textbf{0.738}^{\pm 0.041}$ \\
\hline
\multicolumn{6}{c}{\textbf{0-shot case - excluded Tumor class from train}} \\
\hline
$\text{SAM2}_{Tuned}$ & $0.708^{\pm 0.055}$ & $0.853^{\pm 0.063}$ & $0.735^{\pm 0.080}$ & $0.704^{\pm 0.077}$ & $0.538^{\pm 0.125}$ \\
SAMSA (ours) & $\textbf{0.760}^{\pm 0.053}$ & $\textbf{0.881}^{\pm 0.048}$ & $\textbf{0.821}^{\pm 0.081}$ & $\textbf{0.763}^{\pm 0.095}$ & $\textbf{0.576}^{\pm 0.072}$ \\
\end{tabular}
\end{table}

As seen in \cref{class_table}, when the tumor class is excluded from training, $\text{SAM2}_{Tuned}$ performance drops by $0.14$, falling below even $\text{SAM2}_{Base}$ performance for tumor detection. Despite this, its overall Macro performance remains significantly better ($+0.185$). Similarly, SAMSA experiences a performance decrease on tumor class ($-0.17$), but crucially maintains the highest  tumor detection capability. Additionally, SAMSA achieves a higher overall Macro result ($+0.052$), suggesting that incorporating spectral information provides meaningful advantages for generalizing to unseen classes.

\begin{table}[b]
  \centering
  \caption{Model performance Macro D@0.5 using cross and mixed training } \label{tab: mix_training}
  \begin{tabular}{l|c|cc|cc|cc}
   Training$\rightarrow$  & \textbf{None} & \multicolumn{2}{c|}{\textbf{Heipor}} & \multicolumn{2}{c|}{\textbf{Hib}} & \multicolumn{2}{c}{\textbf{Mixed}} \\
   \cmidrule(lr){1-8} 
   HSI Channels  & - & \multicolumn{2}{c|}{100} & \multicolumn{2}{c|}{128} & \multicolumn{2}{c}{238} \\
    \midrule
    Num Classes  & - & \multicolumn{2}{c|}{20} & \multicolumn{2}{c|}{4} & \multicolumn{2}{c}{24} \\
    \midrule
   \multirow{2}{*}{Test$\downarrow$} & $SAM2_{Base}$ & SAM2 & \multirow{2}{*}{SAMSA} & $SAM2$ & \multirow{2}{*}{SAMSA} & $SAM2$ & \multirow{2}{*}{SAMSA} \\
     &  (0-shot) & $_{Tuned}$ & & $_{Tuned}$ & & $_{Tuned}$& \\
    \midrule
    \textbf{Heipor} & 0.600 & 0.806 & 0.811 & 0.445 & 0.433 & 0.807 & 0.810 \\
    \textbf{Hib} & 0.523 & 0.454 & 0.497 & 0.771 & 0.810 & 0.695 & 0.765 \\
  \end{tabular}
  \label{tab:cross_dataset_performance}
\end{table}

Secondly, our approach uniquely enables training across datasets with different spectral properties by collapsing spectral information to a single channel regardless of band count or resolution. In \cref{tab: mix_training}, cross-dataset generalization (training on one dataset, testing on another) performs poorly even below the zero-shot $\text{SAM2}_{Base}$ baseline. However, mixed training significantly improves results. While $\text{SAM2}_{Tuned}$ shows inconsistent benefits from mixed training (improved on Heipor, decreased on Hib), SAMSA maintains balanced performance, outperforming $\text{SAM2}_{Tuned}$ on both datasets (Hib $+0.07$, Heipor $+0.003$). This confirms SAMSA's ability to generalize across heterogeneous HSI datasets with varying spectral properties and clinical domains.



In \cref{example_outputs} we present qualitative results on the Hib dataset. When clicking on vascular tissue (a), $\text{SAM2}_{Tuned}$ (d) struggles to effectively segment the vascular class without spectral information. The SA map (e) clearly identifies vascular structures but introduces noise around the tumor region. In contrast, SAMSA (f) produces a well-localized probability map for vascular tissue. For the Heipor dataset, clicking on small bowel tissue (g) demonstrates SAMSA's ability to precisely delineate class boundaries compared to the ground truth (h).

\section{Conclusion}



SAMSA is a unique method for generalizing across different HSI datasets, enabling effective segmentation in scenarios with limited training data and diverse imaging conditions. The proposed framework's ability to combine spectral and RGB information provides significant advantages, particularly in detecting challenging medical structures and maintaining performance across different datasets. Our approach shows promise in handling unseen classes and adapting to heterogeneous HSI datasets under low data regimes, opening new possibilities for flexible and robust hyperspectral interactive medical image analysis.

\newpage
\bibliographystyle{splncs04}
\bibliography{mybibliography}

\end{document}